\documentclass[10pt,twocolumn,letterpaper]{article}
\usepackage[final]{cvpr} 
\usepackage{graphicx}
\usepackage{amsmath}
\usepackage{amssymb}
\usepackage{booktabs}
\usepackage{multirow}
\usepackage{booktabs}
\usepackage[T1]{fontenc}
\usepackage[pagebackref,breaklinks,colorlinks]{hyperref}

\usepackage[capitalize]{cleveref}
\crefname{section}{Sec.}{Secs.}
\Crefname{section}{Section}{Sections}
\Crefname{table}{Table}{Tables}
\crefname{table}{Tab.}{Tabs.}

\def\cvprPaperID{1029} 
\def\confName{CVPR}
\def\confYear{2023}

\begin{document}

\title{KD-DLGAN: Data Limited Image Generation via Knowledge Distillation}

\author {
    Kaiwen Cui\textsuperscript{\rm 1}, 
    Yingchen Yu\textsuperscript{\rm 1},
    Fangneng Zhan\textsuperscript{\rm 2},
    Shengcai Liao\textsuperscript{\rm 3}, 
    Shijian Lu\textsuperscript{\rm 1}\thanks{corresponding author.}, 
    Eric Xing\textsuperscript{\rm 4} \\
    \textsuperscript{\rm 1} Nanyang Technological University, 
    \textsuperscript{\rm 2} Max Planck Institute for Informatics\\
    \textsuperscript{\rm 3} Inception Institute of Artificial Intelligence\\
    \textsuperscript{\rm 4} Mohamed bin Zayed
    University of Artificial Intelligence\\
    {\tt\small \{Kaiwen.Cui, Yingchen.Yu, Shijian.Lu\}@ntu.edu.sg, fzhan@mpi-inf.mpg.de} \\
    {\tt\small shengcai.liao@inceptioniai.org, Eric.Xing@mbzuai.ac.ae}\\
}

\maketitle

\begin{abstract}
Generative Adversarial Networks (GANs) rely heavily on large-scale training data for training high-quality image generation models. With limited training data, the GAN discriminator often suffers from severe overfitting which directly leads to degraded generation especially in generation diversity. Inspired by the recent advances in knowledge distillation (KD), we propose KD-DLGAN, a knowledge-distillation based generation framework that introduces pre-trained vision-language models for training effective data-limited generation models. KD-DLGAN consists of two innovative designs. The first is aggregated generative KD that mitigates the discriminator overfitting by challenging the discriminator with harder learning tasks and distilling more generalizable knowledge from the pre-trained models. The second is correlated generative KD that improves the generation diversity by distilling and preserving the diverse image-text correlation within the pre-trained models. Extensive experiments over multiple benchmarks show that KD-DLGAN achieves superior image generation with limited training data. In addition, KD-DLGAN complements the state-of-the-art with consistent and substantial performance gains. Note that codes will be released.
\end{abstract}

\section{Introduction}  \label{sec:1}

\begin{figure}[t!] 
\begin{center}
   \includegraphics[width=1\linewidth]{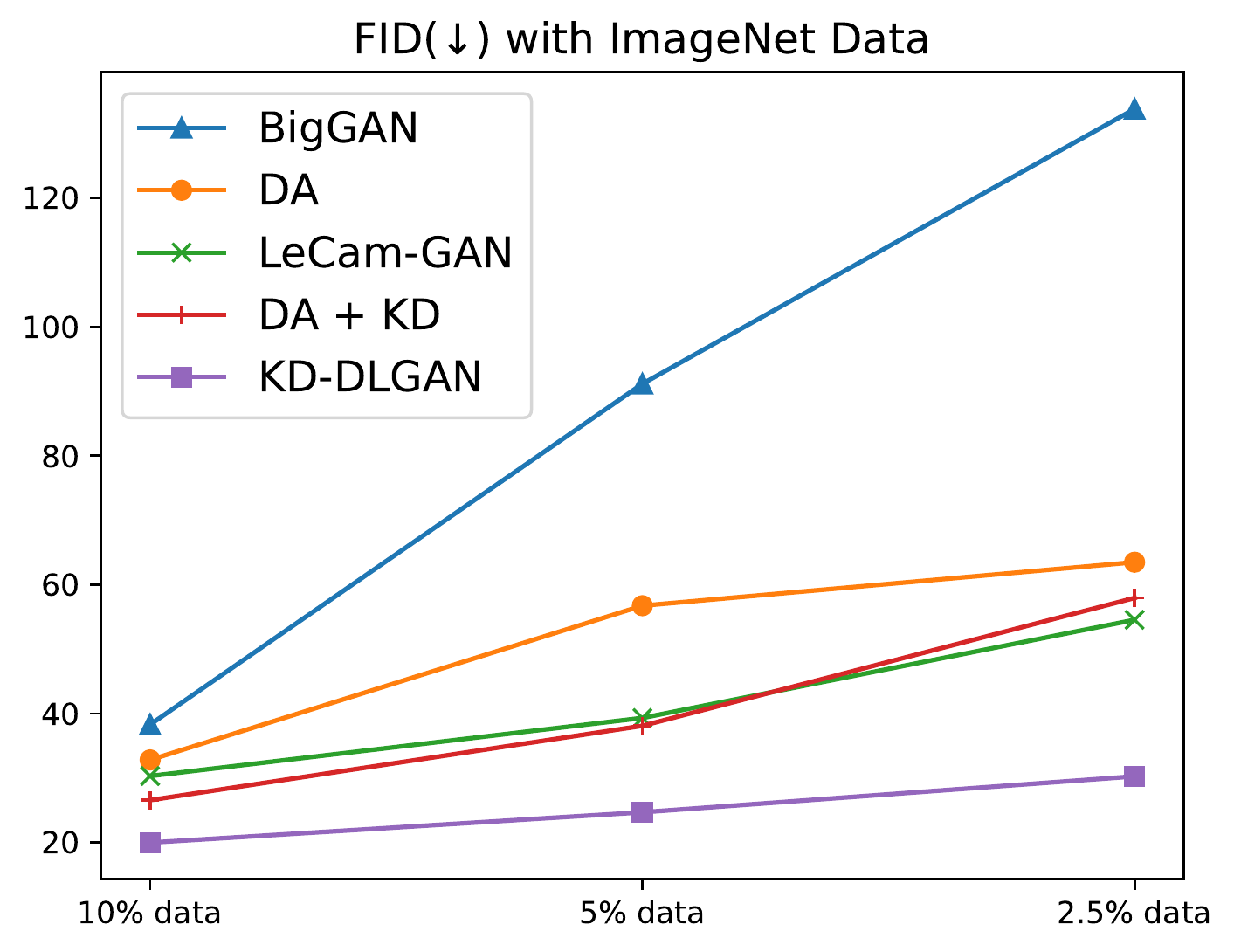} 
\end{center}
  \caption{ 
With limited training samples, state-of-the-art GANs such as BigGAN suffer from clear discriminator overfitting which directly leads to degraded generation. The recent work attempts to mitigate the overfitting via mass data augmentation in DA~\cite{zhao2020differentiable} or regularization in LeCam-GAN~\cite{tseng2021regularizing}. The proposed KD-DLGAN distills the rich and diverse text-image knowledge from the powerful visual-language model to the discriminator which greatly mitigates the discriminator overfitting. Additionally, KD-DLGAN is designed specifically for image generation tasks, which also outperforms the vanilla knowledge distillation (DA+KD) greatly.}
\label{fig:1}
\end{figure}

Generative Adversarial Networks (GANs)~\cite{goodfellow2014generative} have become the cornerstone technique in various image generation tasks. On the other hand, effective training of GANs relies heavily on large-scale training images that are usually laborious 
and expensive 
to collect. With limited training data, the discriminator in GANs often suffers from severe overfitting~\cite{zhao2020differentiable,tseng2021regularizing}, leading to degraded generation as shown in Fig.~\ref{fig:1}. Recent works attempt to address this issue from two major perspectives: i) massive data augmentation that aims to expand the distribution of the limited training data~\cite{zhao2020differentiable}; ii) model regularization that introduces regularizers to modulate the discriminator learning~\cite{tseng2021regularizing}. We intend to mitigate the discriminator overfitting from a new perspective.

Recent studies show that knowledge distillation (KD) from powerful vision-language models such as CLIP~\cite{radford2021learning} can effectively relieve network overfitting in  visual recognition tasks~\cite{Cheng_2021_CVPR,Andonian_2022_CVPR,Ma_2022_CVPR,Wang_2022_CVPR}.
Inspired by these prior studies, we explore KD for data-limited image generation, aiming to mitigate the discriminator overfitting by distilling the rich image-text knowledge from vision-language models. One intuitive approach is to adopt existing KD methods~\cite{romero2014fitnets,hinton2015distilling} for training data-limited GANs, e.g., by forcing the discriminator to mimic the representation space of vision-language models. However, such approach does not work well as most existing KD methods are designed for visual recognition instead of GANs as illustrated in \textit{DA+KD} in Fig.~\ref{fig:1}.

We propose KD-DLGAN, a knowledge-distillation based image generation framework that introduces the idea of generative KD for training effective data-limited GANs. KD-DLGAN is designed based on two observations in data-limited image generation: 1) the overfitting is largely attributed to the simplicity of the discriminator task, i.e., the discriminator can easily memorize the limited training samples and distinguish them with little efforts; 2) the degradation in data-limited generation is largely attributed to poor generation diversity, i.e., the trained data-limited GAN models tend to generate similar images.

Inspired by the two observations, we design two generative KD techniques that jointly distill knowledge from CLIP~\cite{radford2021learning} to the GAN discriminator for effective training of data-limited GANs. The first is aggregated generative KD (AGKD) that challenges the discriminator by 
forcing fake samples to be similar to real samples 
while mimicking CLIP's visual feature space. It mitigates the discriminator overfitting by aggregating features of real and fake samples
and distilling generalizable CLIP knowledge concurrently. The second is correlated generative KD (CGKD) that strives to distill CLIP image-text correlations to the GAN discriminator. It improves the generation diversity by enforcing the diverse correlations between images and texts, ultimately improving the generation performance. The two designs distill the rich yet diverse CLIP knowledge which effectively mitigates the discriminator overfitting and improve the generation as illustrated in \textit{KD-DLGAN} in Fig.~\ref{fig:1}.

The main contributions of this work can be summarized in three aspects. \textit{First,} we propose KD-DLGAN, a novel image generation framework that introduces knowledge distillation for effective GAN training with limited training data. To the best of our knowledge, this is the first work that exploits the idea of knowledge distillation in data-limited image generation. \textit{Second,} we design two generative KD techniques including aggregated generative KD and correlated generative KD that mitigate the discriminator overfitting and improves the generation performance effectively. \textit{Third,}  extensive experiments over multiple widely adopted benchmarks show that KD-DLGAN achieves superior image generation and it also complements the state-of-the-art with consistent and substantial performance gains.

\section{Related Works}    \label{sec:2}

\textbf{Generative Adversarial Network:} 
Generative adversarial network~\cite{goodfellow2014generative} (GAN) has achieved significant progress in automated image generation and editing~\cite{koksal2020rf,zhan2021unbalanced, nazeri2019edgeconnect,yu2022towards}. Following the idea in~\cite{goodfellow2014generative}, quite a few generation applications have been developed in the past few years. They intend to generate more realistic images from different aspects by adopting novel training objectives~\cite{arjovsky2017wasserstein, gulrajani2017improved, mao2017least}, designing more advanced networks~\cite{miyato2018spectral, miyato2018cgans, zhang2019self}, introducing elaborately designed training strategies~\cite{karras2017progressive,zhang2017stackgan,liu2020diverse}, etc. On the contrary,  most existing GANs rely heavily on large-scale training samples.
With only limited samples, they often suffer from clear discriminator overfitting and severe  generation degradation. 

We target data-limited image generation, which intends to train effective GAN models with limited number of samples yet without sacrificing much generation performance. 

\textbf{Data-Limited Image Generation:} Data-limited image generation is a challenging yet meaningful task for circumventing the laborious image collection process. Prior studies~\cite{webster2019detecting, gulrajani2020towards} suggest that one of the main obstacles of training effective data-limited GAN lies with the overfitting of GAN discriminator. 
Recent studies~\cite{zhao2020differentiable,karras2020training,tseng2021regularizing,jiang2021deceive,yang2021data,cui2022genco,huangmasked} attempt to mitigate the overfitting issue mainly through massive data augmentation or model regularization. For example, \cite{zhao2020differentiable} introduces different types of differentiable augmentation to
improve the generation performance. \cite{karras2020training} presents an adaptive augmentation mechanism that prevents undesirable leaking of augmentation to the generated images. \cite{tseng2021regularizing} introduces a regularization scheme to modulate the discriminator. Several studies instead employ external knowledge for data-limited image generation. For example, \cite{mo2020freeze} pretrains the GAN model on a larger dataset. \cite{Kumari_2022_CVPR} employs off-the-shelf models as additional discriminators to improve the data-limited GAN performance.

We target to tackle the discriminator overfitting issue from a new perspective of knowledge distillation and design two generative knowledge distillation techniques to effectively distill knowledge from a powerful vision-language model to the GAN discriminator.

\begin{figure*}[t!] 
\begin{center}
   \includegraphics[width=0.9\linewidth]{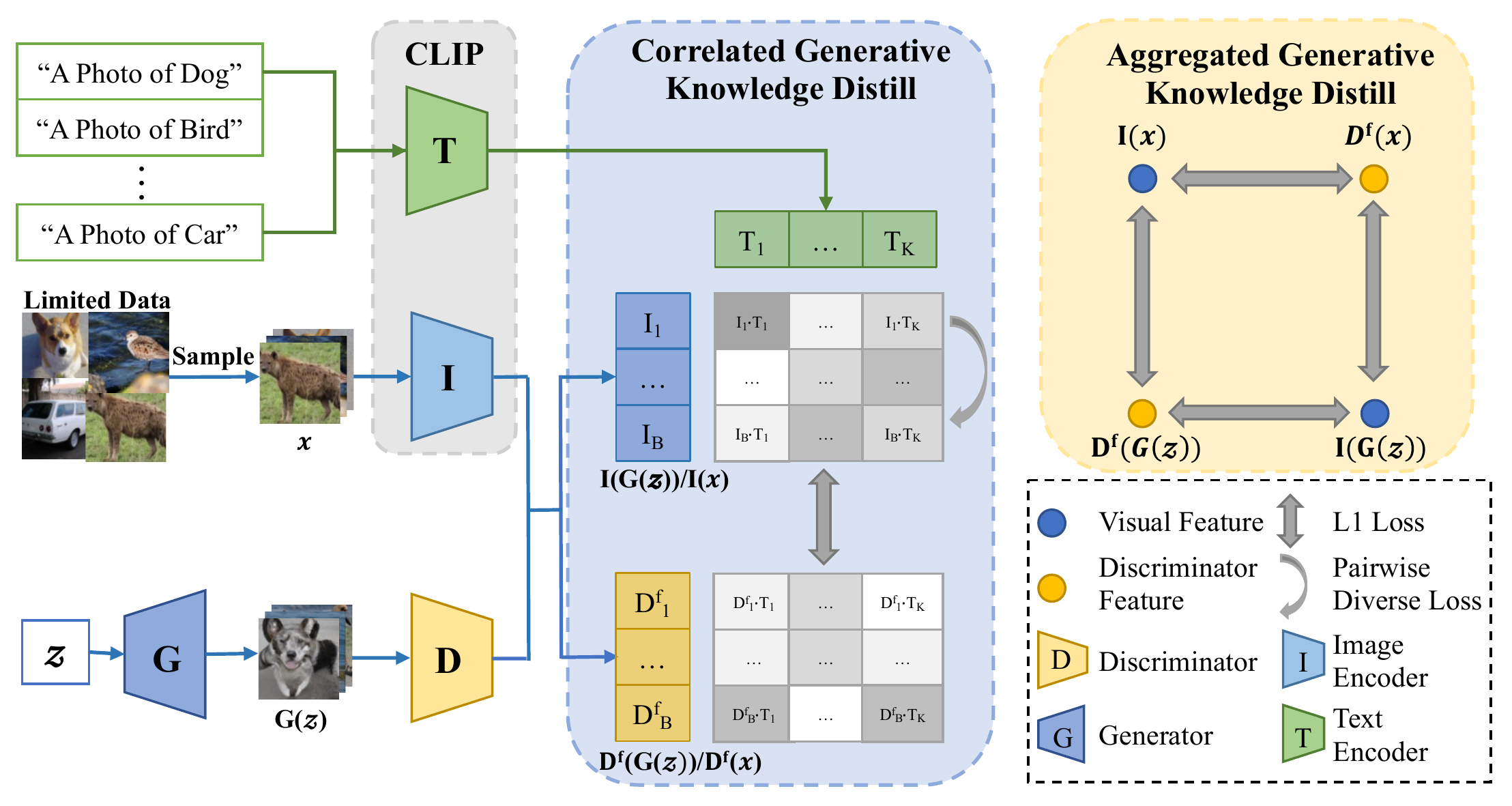}
\end{center}
   \caption{Architecture of the proposed KD-DLGAN: KD-DLGAN distills knowledge from the powerful vision-language model CLIP~\cite{radford2021learning} to the discriminator for effective GAN training with limited training data. We design two 
   generative knowledge distillation techniques that work orthogonally, including aggregated generative knowledge distillation and correlated generative knowledge distillation.  Aggregated generative knowledge distillation mitigates the discriminator overfitting with harder learning tasks and distills general knowledge from the pretrained model.  Correlated generative knowledge distillation improves the generation diversity by distilling and preserving the diverse image-text correlation within the pretrained model. 
   }
\label{fig:model}
\end{figure*}

\textbf{Knowledge Distillation:}
Knowledge distillation is a general technique for supervising the training of student networks by transferring the knowledge of trained teacher networks. Knowledge distillation is initially designed for model compression~\cite{bucilua2006model} 
via mimicking
the output of an ensemble of models. 
\cite{ba2014deep} 
further compresses deep networks into shallower but wider ones via mimicking the logits.
\cite{hinton2015distilling} 
presents a more general knowledge distillation technique by applying the prediction
of the teacher model as a soft label. \cite{tung2019similarity} measures the similarity between pairs of instances in the teacher’s feature space and encourages the student to mimic the pairwise similarity. 
Leveraging on these ideas, 
Knowledge distillation has recently been widely explored and adopted in various applications such as image classification~\cite{yun2020regularizing,zhu2018knowledge}, domain adaptation~\cite{gupta2016cross,nguyen2021unsupervised} , object detection~\cite{chen2017learning,chawla2021data}, semantic segmentation~\cite{liu2019structured,yang2022cross}, etc.

We introduce knowledge distillation into the data-limited image generation task for mitigating its overfitting issue. To the best of our knowledge, this is the first work that explores knowledge distillation in  data-limited image generation.

\section{Method}

\subsection{Overview}  

We describes the detailed methodology of the proposed KD-DLGAN in this section.  As shown in Fig. \ref{fig:model}, we introduce knowledge distillation for data-limited image generation. 
Specifically, leveraging CLIP~\cite{radford2021learning} as the teacher model, we design two generative knowledge distillation techniques, including aggregated generative knowledge
distillation that leads to less distinguishable real-fake samples for the discriminator while distilling more generalizable knowledge from the pretrained model, and correlated generative knowledge distillation that encourages the discriminator to mimic the diverse vision-language correlation in CLIP. 
The ensuing subsections will describe the problem definition of data-limited image generation, details of the proposed aggregated generative knowledge
distillation and correlated generative knowledge distillation, and the overall training objective, respectively.

\subsection{Problem Definition}
\label{sec:Problem Definition}
GANs greatly change the paradigm of image generation. 
Each GAN consists of a discriminator $D$ and a generator $G$. The general loss function for discriminator and generator is formulated as:
\begin{equation}
\label{eqn:general D}
\mathcal{L}_{d}(D; x, G(z))  = \mathbb{E}[\log(D(x))]  
  +  \mathbb{E}[\log(1-D(G(z))]
\end{equation}
\begin{equation}
\label{eqn: general G}
\mathcal{L}_{g}(D; G(z)) = \mathbb{E}[\log(1-D(G(z))]
\end{equation}

\noindent where $\mathcal{L}_{d}$ and $\mathcal{L}_{g}$  denote the general discriminator loss and generator loss, respectively. $x$ denotes a training sample and $z$ is randomly sampled from Gaussian Distribution. 

In data-limited image generation, the discriminator in GANs tends to memorize the exact training data and is prone to suffer from overfitting, leading to sub-optimal image generation. Recent studies show that knowledge distillation from powerful and generalizable models can relieve overfitting~\cite{Cheng_2021_CVPR,Andonian_2022_CVPR,Ma_2022_CVPR,Wang_2022_CVPR} effectively. 
However, these state-of-the-art knowledge distillation methods are mainly designed for visual recognition tasks, which cannot be naively adapted to the image generation tasks. We design two novel knowledge distillation techniques that can 
greatly improve data-limited image generation, more details to be presented in the following subsections.

\subsection{Aggregated Generative Knowledge Distillation}

We design aggregated generative knowledge distillation~(AGKD) to mitigate the discriminator overfitting  by distilling more generalizable knowledge from the pretrained CLIP~\cite{radford2021learning} model. Thus, we force the discriminator feature space to mimic CLIP visual feature space. Specifically, for real samples $x$, we distill knowledge from CLIP visual feature of $x$ (denoted by $I(x)$) to the discriminator feature of $x$ (last layer feature, which is denoted by $D^{f}(x)$) with L1 loss. Similarly, for generated samples $G(z)$, we also distill  knowledge from CLIP visual feature $I(G(z))$ to the discriminator feature $D^{f}(G(z))$. The knowledge distillation loss $\mathcal{L}^{KD}_{AGKD}$ in AKGD can thus be formulated by:
\begin{align*}\left.\begin{aligned}
\mathcal{L}^{KD}_{AGKD} = |I(x) - D^{f}(x)| + |I(G(z)) - D^{f}(G(z))|
\end{aligned}\right.\end{align*}

AGKD also mitigates the discriminator overfitting by aggregating features of real and fake samples to challenge the discriminator learning. Specifically, we match the CLIP visual feature of real samples $I(x)$ and discriminator features of fake samples $D^{f}(G(z))$, as well as CLIP visual feature of fake samples $I(G(z))$ and discriminator features of real samples $D^{f}(x$) by the L1 loss.  Such  design lowers the real-fake discriminability and makes it harder to distinguish real-fake samples for the discriminator. The aggregated loss $\mathcal{L}^{AGG}_{AGKD}$ can be formulated by:
\begin{align*}\left.\begin{aligned}
\mathcal{L}^{AGG}_{AGKD} = |I(x) - D^{f}(G(z))| + |I(G(z)) - D^{f}(x)| 
\end{aligned}\right.\end{align*}

For effective GAN training, the designed aggregated loss $\mathcal{L}^{AGG}_{AGKD}$  is controlled by a hyper-parameter $p$, where the loss is applied with probability $p$ or skipped with probability 1-$p$.  We empirically set $p$ at 0.7 in our trained networks. Thus, the aggregated loss $\mathcal{L'}^{AGG}_{AGKD}$ can be re-formulated by:
\begin{align*}\left.\begin{aligned}
\mathcal{L'}^{AGG}_{AGKD} = 
\left\{
\setlength{\arraycolsep}{2pt}
\begin{array}{llllll}
\mathcal{L}^{AGG}_{AGKD}, && {\rm if} & q \leq p, \\
0, && {\rm if} & q > p, \\
\end{array}
\right.
\end{aligned}\right.\end{align*}
where $q$ is a random number sampled between 0 and 1. 

The overall AGKD loss $\mathcal{L}_{AGKD}$ can be formulated by: 
\begin{align}\left.\begin{aligned}
\mathcal{L}_{AGKD} =  \mathcal{L}^{KD}_{AGKD}  + \mathcal{L'}^{AGG}_{AGKD}
\label{eqn:AGKD}
\end{aligned}\right.\end{align}

\subsection{Correlated Generative Knowledge Distillation}

Correlated generative knowledge distillation (CGKD) aims to improve the generation diversity by two steps:
1) it enforces the diverse correlations between generated images and texts in CLIP with a pairwise diversity loss ($\mathcal{L}^{PD}_{CGKD}$); 2) it distills the diverse correlations from CLIP to the GAN discriminator with a distillation loss ($\mathcal{L}^{KD}_{CGKD}$). 

To achieve diverse image-text correlations, it first builds the correlations (indicated by inner products) between CLIP visual features of generated images $I(G(z))\in \mathbb{R}^{B \times M}$ and CLIP texts features $T \in \mathbb{R}^{K \times M}$. Here, $B$ is the batch size of generated images, $K$ is the number of texts and $M$ is the features dimension for each text feature or each image feature.
For conditional datasets and unconditional datasets, we employ the corresponding image labels as input texts and predefine a set of relevant text labels as input texts, respectively. 
Details of text selection for our datasets are introduced in the supplementary material. Thus, the correlation $C_T \in \mathbb{R}^{B \times K}$ between $I(G(z))$ and $T$ (i.e., their L2-normalized inner products) can be defined as follows:
\begin{equation*}
C_T = \frac{I(G(z)) \cdot T'}{||I(G(z)) \cdot T'||_2},
\end{equation*}
where $T'$ is the transpose of $T$. 

With the defined correlation, the diverse CLIP image-text correlations can be extracted in a pairwise manner.
Specifically, for each image-text correlation $C_T[i,:] \in \mathbb{R}^{K}$, we diversify it with another image-text correlation $C_T[j,:] \in \mathbb{R}^{K}$ by minimizing the cosine similarity between them. Note  [$i$,:] or [$j$,:]  denotes the $i$-th or $j$-th row in $C_T$ and $j\neq i$. The pairwise diversity loss $\mathcal{L}^{PD}_{CGKD}$ can thus be defined as the sum of the cosine similarity of all pairs:
\begin{align*}\left.\begin{aligned}
\mathcal{L}^{PD}_{CGKD} = \sum_{i=1}^{K}\sum_{j=1, j\neq i}^{K} Cos(C_T[i,:] , C_T[j,:]),
\end{aligned}\right.\end{align*}
where $Cos(\overrightarrow a,\overrightarrow b)$ indicates the cosine similarity between the two vectors $\overrightarrow a$ and $\overrightarrow b$.

Then, the obtained diverse correlations are distilled from CLIP to the GAN discriminator, aiming to improve the generation diversity. We build the correlations $C_S \in \mathbb{R}^{B \times K} $ between discriminator features of generated samples $D^f(G(z)) \in \mathbb{R}^{B \times M}$ and CLIP text features $T \in \mathbb{R}^{K \times M} $ as follows:
\begin{align*}\left.\begin{aligned}
C_S = \frac{D^f(G(z))\cdot T'}{||D^f(G(z))\cdot T'||_2}
\end{aligned}\right.\end{align*}
The correlation distillation from $C_T$ to $C_S$ is defined by the L1 loss between them:
\begin{align*}\left.\begin{aligned}
\mathcal{L}^{KD}_{CGKD} = |C_T  - C_S|
\end{aligned}\right.\end{align*}
The overall CGKD loss $\mathcal{L}_{CGKD}$ can thus be defined by: 
\begin{align}\left.\begin{aligned}
\mathcal{L}_{CGKD} =  \mathcal{L}^{PD}_{CGKD} + \mathcal{L}^{KD}_{CGKD}  
\label{eqn:CGKD}
\end{aligned}\right.\end{align}

\begin{table*}
\begin{center}
\small 
\renewcommand\tabcolsep{14pt}
\resizebox{0.85\linewidth}{!}{
\begin{tabular}{lccccc}
\toprule
 \multirow{2}*{{Methods}}  & \multicolumn{3}{c}{100-shot} & \multicolumn{2}{c}{AFHQ} \\
 \cmidrule(lr){2-4}  \cmidrule(lr){5-6}
 & Obama & Grumpy Cat & Panda & Cat & Dog \\
\midrule 
DA~\cite{zhao2020differentiable} + KD (CLIP~\cite{radford2021learning}) &45.22&25.62& 11.24 & 38.31 & 55.13\\
\midrule 
DA~\cite{zhao2020differentiable} (Baseline)  &46.87& 27.08& 12.06 & {42.44}  & {58.85} \\
+ \textbf{KD-DLGAN (Ours)}   & \textbf{31.54} \textpm \ 0.27 & {\textbf{20.13}} \textpm \ 0.13 & {\textbf{8.93}} \textpm \ 0.06 & \textbf{32.99} \textpm \ 0.10  & \textbf{51.63} \textpm \ 0.17\\
\midrule 
LeCam-GAN~\cite{tseng2021regularizing}  & 33.16 & 24.93 & 10.16 &{34.18}  & {54.88}\\
+ \textbf{KD-DLGAN (Ours)}  &  \textbf{29.38}  \textpm \ 0.15 & \textbf{19.65} \textpm \ 0.17 &\textbf{ 8.41 }\textpm \ 0.05 &  \textbf{31.89} \textpm \ 0.09 & \textbf{50.22} \textpm \ 0.16 \\
\midrule 
InsGen~\cite{yang2021data}   & 45.85 & 27.48 & 12.13 &41.33 & 55.12 \\
+ \textbf{KD-DLGAN (Ours)}  &\textbf{38.28} \textpm \ 0.25 & \textbf{22.16} \textpm \ 0.12 & \textbf{9.51} \textpm \ 0.07 & \textbf{32.39} \textpm \ 0.08  & \textbf{50.13} \textpm \ 0.12\\
\midrule
APA~\cite{jiang2021deceive}   & 43.75 & 28.49 &  12.34  & 39.13 & 54.15\\
+ \textbf{KD-DLGAN (Ours)}  &{\textbf{34.68}} \textpm \ 0.21 &\textbf{23.14} \textpm \ 0.14 & {\textbf{8.70}} \textpm \ 0.05 & \textbf{31.77} \textpm \ 0.09 & \textbf{51.23} \textpm \ 0.13\\
\midrule
ADA~\cite{karras2020training} & 45.69 & 26.62& 12.90& 40.77 & 56.83\\
+ \textbf{KD-DLGAN (Ours)}  & {\textbf{31.78}} \textpm \ 0.22 & {\textbf{19.76}} \textpm \ 0.11 & {\textbf{8.85}} \textpm \ 0.05 & \textbf{32.81} \textpm \ 0.10  & \textbf{51.12} \textpm \ 0.15\\ 
\bottomrule
\end{tabular}
}
\end{center}
\caption{Comparison with the state-of-the-art over 100-shot and AFHQ: KD-DLGAN outperforms and complements the state-of-the-art data-limited image generation approaches consistently. 
In addition, KD-DLGAN outperforms vanilla knowledge
distillation in DA + KD (CLIP~\cite{radford2021learning}) consistently.  All the compared methods employ StyleGAN-v2~\cite{karras2020analyzing} as backbone. We report FID($\downarrow$) averaged over three runs.
}
\label{Low Shot Generation}
\end{table*}

\begin{figure*}[t] 
\begin{center}
\includegraphics[width=0.9\linewidth]{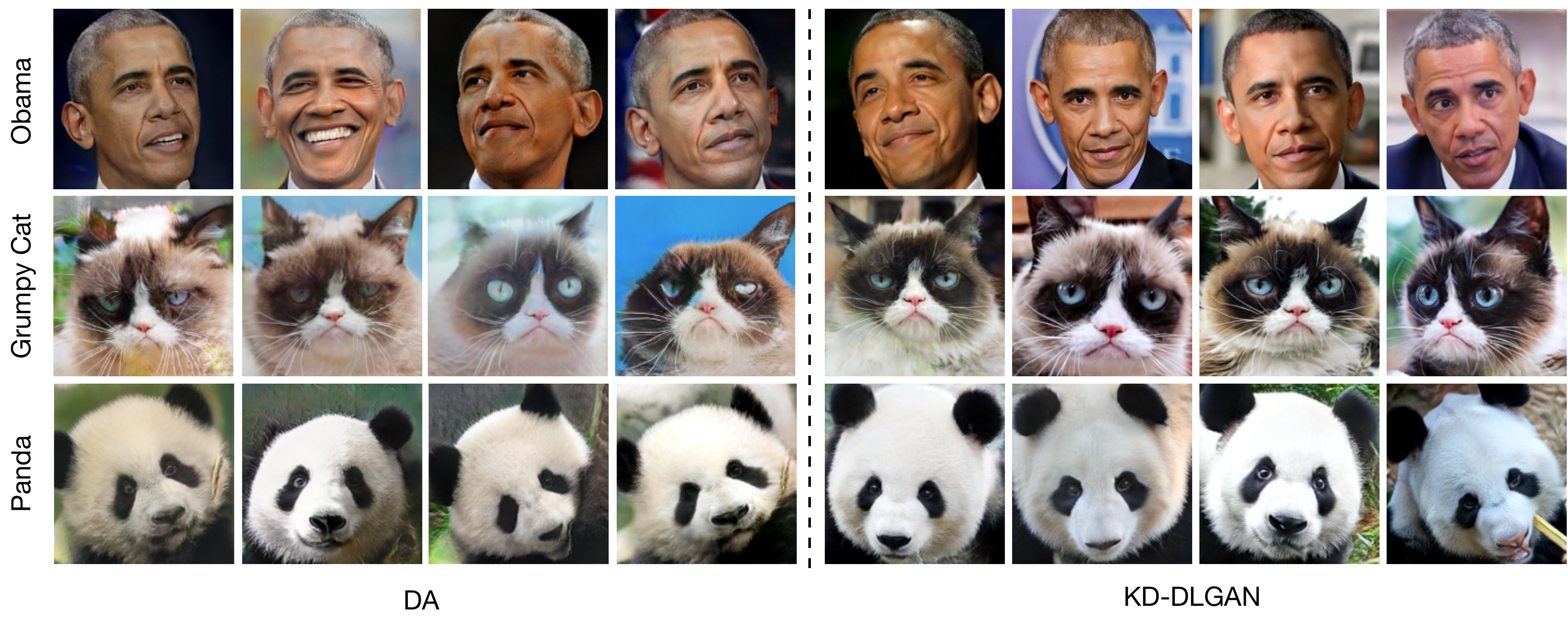}
\end{center}
   \caption{
    Qualitative comparison with the state-of-the-art over 100-shot: Samples generated by KD-DLGAN are clearly more realistic than those generated by DA~\cite{zhao2020differentiable}, the state-of-the-art data-limited generation approach.
   }   
\vspace{-2mm}     
\label{fig:qualitative}
\end{figure*}

\subsection{Overall Training Objective} \label{sec:Overall Training Objective}

The overall training objective of the proposed KD-DLGAN can thus be formulated by:
\begin{align}\left.\begin{aligned}
\mathop{min}\limits_{G}\mathop{max}\limits_{D}  \mathcal{L}_{G} + \mathcal{L}_{D}
\end{aligned}\right.\end{align}
where $\mathcal{L}_{G}=\mathcal{L}_{g}$ as introduced in Eq.~\ref{eqn: general G} and $\mathcal{L}_{D}=\mathcal{L}_{d} + \mathcal{L}_{AGKD} + \mathcal{L}_{CGKD}$ as introduced in Eqs.~\ref{eqn:general D}, \ref{eqn:AGKD} and \ref{eqn:CGKD}.

\begin{table*}[t]
\begin{center}
\resizebox{0.85\linewidth}{!}{
\begin{tabular}{lcccccc}
\toprule
\multirow{2}*{Method} & \multicolumn{3}{c}{CIFAR-10} & \multicolumn{3}{c}{CIFAR-100}\\
\cmidrule(lr){2-4} \cmidrule(lr){5-7}
&  \multicolumn{1}{c}{100\% Data}  & \multicolumn{1}{c}{20\% Data}   & \multicolumn{1}{c}{10\% Data} &  \multicolumn{1}{c}{100\% Data}  & \multicolumn{1}{c}{20\% Data}   & \multicolumn{1}{c}{10\% Data} \\
\midrule
DA~\cite{zhao2020differentiable}  + KD (CLIP~\cite{radford2021learning}) & 8.70 \textpm \ 0.02 &13.70 \textpm \ 0.08 &22.03 \textpm \ 0.07 &11.74 \textpm \ 0.02 &21.76 \textpm \ 0.06 &33.93 \textpm \ 0.09 \\
\midrule
DA~\cite{zhao2020differentiable} (Baseline)   & 8.75 \textpm \ 0.03 &  14.53 \textpm \ 0.10& 23.34 \textpm \ 0.09 & 11.99 \textpm \ 0.02&22.55 \textpm \ 0.06&35.39 \textpm \ 0.08  \\
+ \textbf{KD-DLGAN (Ours)} &\textbf{8.42} \textpm \ 0.01 &\textbf{11.01} \textpm \ 0.07  & \textbf{14.20 }\textpm \ 0.06 &\textbf{10.28} \textpm \ 0.03& \textbf{15.60} \textpm \ 0.08 & \textbf{18.03} \textpm \ 0.11  \\
\midrule
APA~\cite{jiang2021deceive} &  8.28 \textpm \ 0.02 & 15.31 \textpm \ 0.04 & 25.98 \textpm \ 0.06 & 11.42 \textpm \ 0.04 & 23.50 \textpm \ 0.06 & 45.79 \textpm \ 0.15 \\ 
+ \textbf{KD-DLGAN (Ours)}    &\textbf{8.26} \textpm \ 0.02 &\textbf{11.15} \textpm \ 0.06  &\textbf{13.86} \textpm \ 0.07 &\textbf{10.23} \textpm \ 0.02& \textbf{19.22} \textpm \ 0.07 & \textbf{27.11} \textpm \ 0.10  \\
\midrule
LeCam-GAN~\cite{tseng2021regularizing}
&{8.46} \textpm \ 0.06 &14.55 \textpm \ 0.08 & 16.69 \textpm \ 0.02 &11.20 \textpm \ 0.09 & 22.45 \textpm \ 0.09 &27.28\textpm \ 0.05 \\
+ \textbf{KD-DLGAN (Ours)}   &\textbf{8.19}  \textpm \ 0.01 & \textbf{11.45} \textpm \ 0.07 & \textbf{13.22}  \textpm \ 0.03 &\textbf{10.12} \textpm \ 0.03 &\textbf{18.70} \textpm \ 0.05 & \textbf{22.40} \textpm \ 0.06 \\
\midrule
ADA~\cite{karras2020training} &8.99  \textpm \ 0.03 &19.87  \textpm \ 0.09 &30.58  \textpm \ 0.11 &12.22  \textpm \ 0.02 & 22.65  \textpm \ 0.10 &27.08  \textpm \ 0.15\\
+ \textbf{KD-DLGAN (Ours)}  &\textbf{8.46} \textpm \ 0.02 &\textbf{ 14.12 }\textpm \ 0.10  &\textbf{16.88} \textpm \ 0.08 &\textbf{10.48} \textpm \ 0.04&\textbf{19.26}  \textpm \ 0.06 &\textbf{20.62} \textpm \ 0.09   \\

\bottomrule
\end{tabular}}
\end{center}
\caption{
Comparison with the state-of-the-art over CIFAR-10 and CIFAR 100: KD-DLGAN outperforms and complements the state-of-the-art clearly.  KD-DLGAN also performs better than vanilla knowledge distillation in DA + KD (CLIP~\cite{radford2021learning}) consistently as well.  All the compared methods employ BigGAN~\cite{brock2018large} as backbone. And we report FID($\downarrow$) averaged over three runs.
}
\label{tab:CIFAR-10-100-biggan}
\end{table*}

\section{Experiments}
In this section, we conduct extensive experiments to evaluate our KD-DLGAN. We first introduce the datasets and the evaluation metrics used in our experiments. 
We then benchmark KD-DLGAN with StyleGAN-v2~\cite{karras2020analyzing} and BigGAN~\cite{brock2018large}.  
Moreover, we conduct extensive ablation studies 
and discussions 
to support our designs. 

\subsection{Datasets and Evaluation Metrics} \label{sec:dataset}
We conduct experiments over  the following datasets: CIFAR~\cite{krizhevsky2009learning}, ImageNet~\cite{deng2009imagenet}, 100-shot~\cite{zhao2020differentiable} and  AFHQ~\cite{si2011learning}. Datasets details are provided in the supplementary material.
We perform evaluations with 
Frechet Inception Distance (FID)~\cite{heusel2017gans} and inception score (IS)~\cite{salimans2016improved}.

\begin{table*}[t]
\begin{center}
\resizebox{0.85\linewidth}{!}{
\begin{tabular}{lcccccc}
\toprule
\multirow{2}*{Method} & \multicolumn{2}{c}{10\% training data} & \multicolumn{2}{c}{5\% training data} & \multicolumn{2}{c}{2.5\% training data} \\
\cmidrule(lr){2-3}
\cmidrule(lr){4-5}
\cmidrule(lr){6-7}
&  IS$\uparrow$  & \multicolumn{1}{c}{FID$\downarrow$}   &  IS$\uparrow$  & \multicolumn{1}{c}{FID$\downarrow$}  &  IS$\uparrow$  & \multicolumn{1}{c}{FID$\downarrow$}\\
DA~\cite{zhao2020differentiable} + KD (CLIP~\cite{radford2021learning}) & 13.29  \textpm \ 0.50 & 26.58  \textpm \ 0.21  & 11.63  \textpm \ 0.29 & 38.11  \textpm \ 0.33 & 9.43  \textpm \ 0.25 & 57.95  \textpm \ 0.41\\
\midrule
DA~\cite{zhao2020differentiable} (Baseline)  & 12.76 \textpm \ 0.34  & 32.82   \textpm \ 0.18 &9.63 \textpm \ 0.21 &56.75  \textpm \ 0.35&8.17 \textpm \ 0.28&63.49  \textpm \ 0.51  \\

+ \textbf{KD-DLGAN (Ours)}  &\textbf{14.25} \textpm \ 0.66&\textbf{19.99 } \textpm \ 0.11 &\textbf{12.71} \textpm \ 0.34 &\textbf{24.70 } \textpm \ 0.14 &\textbf{13.45} \textpm \ 0.51& \textbf{30.27}  \textpm \ 0.16\\
\midrule
LeCam-GAN~\cite{tseng2021regularizing} &11.59 \textpm  \ 0.44 & 30.32 \textpm  \ 0.24 &10.53  \textpm  \ 0.22  & 39.33 \textpm  \ 0.27 & 9.99 \textpm  \ 0.26 & 54.55 \textpm  \ 0.46\\
+ \textbf{KD-DLGAN (Ours)}   &\textbf{13.98} \textpm  \ 0.23 &\textbf{ 22.12} \textpm  \ 0.12 &\textbf{13.86} \textpm  \ 0.45 &\textbf{23.85} \textpm  \ 0.21 &\textbf{13.22} \textpm  \ 0.44 & \textbf{31.33} \textpm  \ 0.15\\
\midrule
ADA & 12.67 \textpm \ 0.31 & 31.89 \textpm \ 0.17 &9.44 \textpm 0.25 &43.21  \textpm \ 0.37 &8.54 \textpm \  0.26 & 56.83  \textpm \ 0.48 \\
+ \textbf{KD-DLGAN (Ours)} & \textbf{14.14} \textpm \ 0.32 & \textbf{20.32} \textpm \ 0.10 & \textbf{14.06} \textpm \ 0.39 & \textbf{22.35} \textpm \ 0.11 & \textbf{14.65 } \textpm \ 0.47 & \textbf{28.79} \textpm \ 0.14 \\ 

\bottomrule 
\end{tabular}
}
\end{center}
\caption{
Comparison with the state-of-the-art over ImageNet~\cite{deng2009imagenet}: KD-DLGAN achieves the best performance consistently and 
complements the state-of-the-art. Besides, KD-DLGAN outperforms vanilla knowledge distillation in DA + KD (CLIP~\cite{radford2021learning}) consistently as well. All the compared methods employ BigGAN~\cite{brock2018large} as backbone. We report IS($\uparrow$) and FID($\downarrow$) averaged over three runs.
}
\label{tab:ImageNet-biggan}
\end{table*}

\subsection{Experiments with StyleGAN-v2}
\label{sec:lowshot}

Table~\ref{Low Shot Generation} shows unconditional image generation results over 100-shot and AFHQ datasets, where we employ StyleGAN-v2~\cite{karras2020analyzing} as the backbone.  
Following the data settings in DA~\cite{zhao2020differentiable}, the models are trained with 100 samples (100-shot Obama, Grumpy Cat, Panda), 160 samples (AFHQ Cat) and 389 samples (AFHQ Dog), respectively.

As Row 3 of Table~\ref{Low Shot Generation} shows, including the proposed KD-DLGAN into DA~\cite{zhao2020differentiable} achieves superior performance across all data settings consistently as compared with DA alone (in Row 2), demonstrating the complementary relation between KD-DLGAN and DA~\cite{zhao2020differentiable}. In addition, the vanilla knowledge distillation in DA + KD (CLIP~\cite{radford2021learning}) (Row 1) trains the GAN discriminator to mimic the visual feature representation of CLIP. We can observe that KD-DLGAN outperforms DA + KD (CLIP~\cite{radford2021learning}) consistently as well, indicating that the performance gain in KD-DLGAN is largely attributed to our generative knowledge distillation designs instead of solely from the powerful vision-language model. Table~\ref{Low Shot Generation} also tabulates the results of KD-DLGAN when implementing over four state-of-the-art data-limited generation approaches including LeCam-GAN~\cite{tseng2021regularizing}, InsGen~\cite{yang2021data}, APA~\cite{jiang2021deceive} and ADA~\cite{karras2020training}. We can see that KD-DLGAN complement all the state-of-the-art consistently, demonstrating the superior generalization and complementary  property of our proposed KD-DLGAN.

\begin{table*}[!h]
\begin{center}
\resizebox{0.75\linewidth}{!}{
\begin{tabular}{lcccccc}
\toprule
\multirow{2}*{Method} & \multirow{2}*{AGKD} & \multirow{2}*{CGKD} & \multicolumn{2}{c}{CIFAR-10} &\multicolumn{2}{c}{100-shot} \\
&&  & \multirow{1}*{20\% data} & \multirow{1}*{10\% data} & Obama & Grumpy Cat\\
\midrule
DA~\cite{zhao2020differentiable} (Baseline) &&&14.53&23.34 & 46.87 & 27.08\\\midrule
&\checkmark&&12.97 \textpm \ 0.08 & 15.85 \textpm \ 0.06& 35.51\textpm \ 0.25  & 23.24 \textpm \ 0.16\\
&&\checkmark&12.77 \textpm \ 0.08 & 18.66 \textpm \ 0.09 & 36.18 \textpm \ 0.22 & 23.17 \textpm \ 0.11\\\midrule
\textbf{Ours} &\checkmark&\checkmark&\textbf{11.01} \textpm \ 0.07  & \textbf{14.20 }\textpm \ 0.06 & \textbf{31.54} \textpm \ 0.27 & \textbf{20.13} \textpm \ 0.13 \\\bottomrule
\end{tabular}
}
\end{center}
\caption{
Quantitative ablation study of KD-DLGAN: AGKD and CGKD in KD-DLGAN both improves the generation performance over the baseline DA~\cite{zhao2020differentiable}. KD-DLGAN performs the best as AGKD and CGKD complement each other. The FIDs ($\downarrow$) are averaged over three runs.
}
\label{tab:ablation}
\end{table*}

Fig. \ref{fig:qualitative} shows qualitative comparison with DA~\cite{zhao2020differentiable}. It can be observed that KD-DLGAN clearly outperforms the state-of-the-art in the data-limited generation, especially in term of the generated shapes and textures.

\subsection{Experiments with BigGAN} \label{sec:cifar}

Table~\ref{tab:CIFAR-10-100-biggan} and Table~\ref{tab:ImageNet-biggan} show the conditional image generation results on CIFAR-10, CIFAR-100 and ImageNet, respectively. All models employ BigGAN~\cite{brock2018large} as the backbone. CIFAR-10 and and CIFAR-100 are trained with 100\% (50K images), 20\% (10K images) or 10\%  (5K images) training data where the FIDs are evaluated over the validation sets (10K images). ImageNet is trained with 10\% (\textasciitilde100K images), 5\% (\textasciitilde 50K images) and 2.5\% (\textasciitilde25K images), where the evaluations are performed over the whole training set (\textasciitilde1.2M images).

The experiments show that our KD-DLGAN outperforms the state-of-the-art substantially. The superior performance is largely attributed to our designed generative knowledge distillation techniques in KD-DLGAN, which mitigates the discriminator overfitting and improves the generation performance effectively. We also show the results of vanilla knowledge distillation from the powerful vision-language model CLIP~\cite{radford2021learning} in Row 1 of Table~\ref{tab:CIFAR-10-100-biggan} and Table~\ref{tab:ImageNet-biggan}. We can see that KD-DLGAN outperforms the vanilla knowledge distillation method by a large margin, indicating that the performance gain is largely attributed to our generative knowledge distillation design instead of the powerful vision-language model. In addition, Table~\ref{tab:CIFAR-10-100-biggan} and Table~\ref{tab:ImageNet-biggan} also show the results of KD-DLGAN when implementing over the state-of-the-art data-limited image generation approaches. KD-DLGAN complements the state-of-the-art and improves the generation performance greatly.

\begin{figure}[t] 
\centering
 \includegraphics[width=0.85\linewidth]{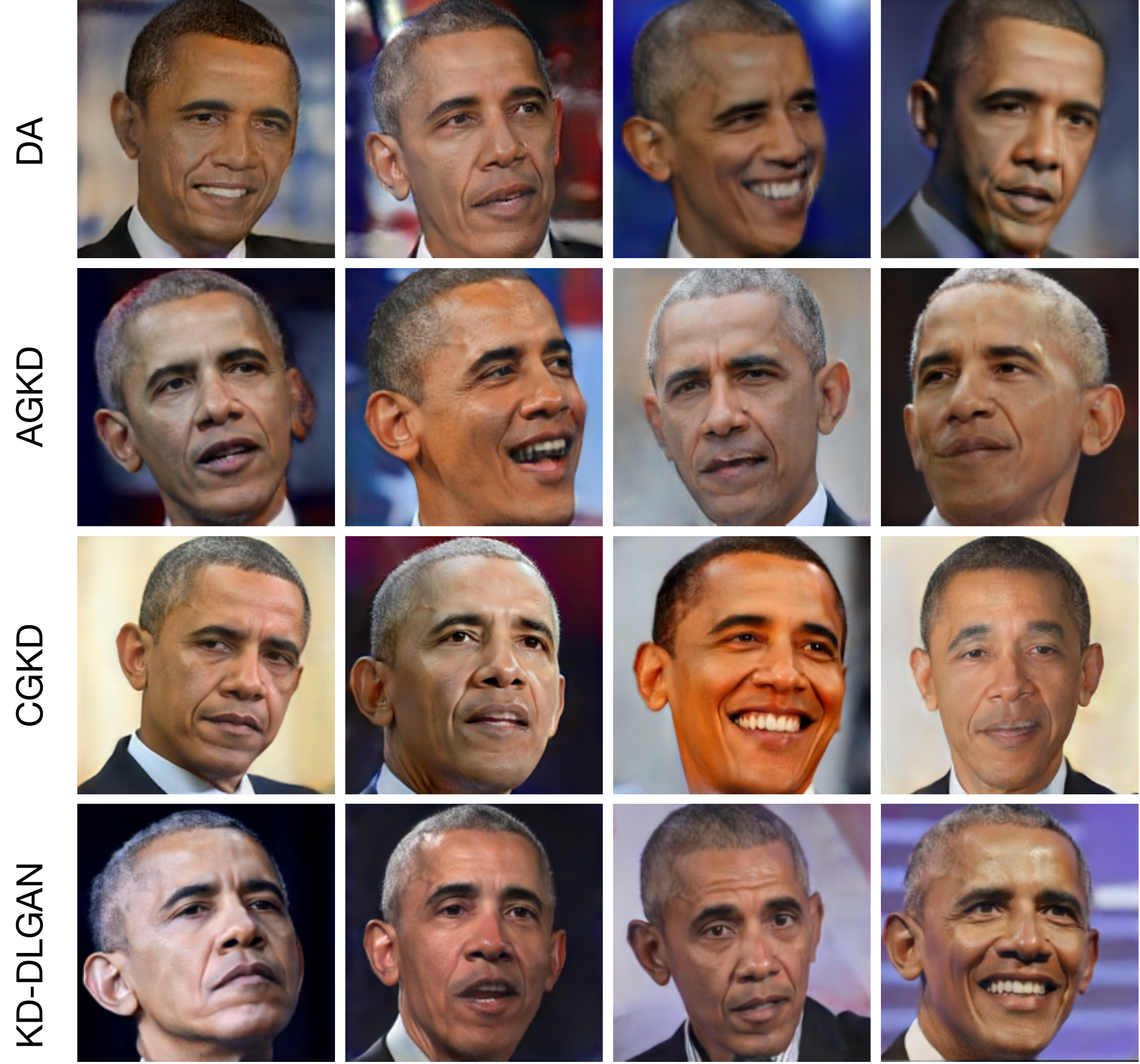}
   \caption{
   Qualitative ablation study over 100-shot Obama: AGKD (Row 2) and CGKD (Row 3) can generate more realistic images than the baseline DA~\cite{zhao2020differentiable} (Row 1), the state-of-the-art in data-limited image generation. KD-DLGAN combining AGKD and CGKD generates the most realistic images. 
   }
\label{abalation-qualitative}
\end{figure}

\subsection{Ablation study} \label{sec:abalation}
The proposed KD-DLGAN consists of two generative knowledge distillation (KD) techniques, namely, AGKD and CGKD. 
The two techniques are  separately evaluated to demonstrate their contributions to the overall generation performance. As Table \ref{tab:ablation} shows, including either AGKD or CGKD clearly outperforms DA~\cite{zhao2020differentiable}, the state-of-the-art in data-limited image generation, demonstrating the effectiveness of the proposed AGKD and CGKD in mitigating the discriminator overfitting and improving the generation performance. In addition, combining AGKD and CGKD leads to the best generation performance which shows that the two KD techniques complement each other.

Qualitative ablation studies in Fig. \ref{abalation-qualitative} show that the proposed AGKD and CGKD can produce clearly more realistic generation than the baseline, demonstrating the effectiveness of these two generative knowledge distillation techniques. In addition, KD-DLGAN combining AGKD and CGKD performs the best, which further verifies that AGKD and CGKD are complementary to each other.

\subsection{Discussion} \label{sec:discussion}
In this subsection, we analyze our KD-DLGAN from several perspectives.
All the experiments are based on the CIFAR-10 and CIFAR-100 dataset with 10\% data.

\textbf{Generation Diversity:} 
The proposed KD-DLGAN improves the generation diversity by enforcing the diverse correlation between images and texts, which eventually improves the generation performance. In this subsection, we evaluate the generation diversity with LPIPS~\cite{zhang2018unreasonable}. Higher LPIPS means better diversity of generated images. As Table~\ref{tab:LPIP} shows, the proposed KD-DLGAN outperforms the baseline DA~\cite{zhao2020differentiable}, the state-of-the-art in data-limited image generation, demonstrating the effectiveness of KD-DLGAN in improving generation diversity. We also show the results of KD-DLGAN without CGKD; it further verifies that the improved generation diversity is largely attributed to the CGKD, which distills diverse image-text correlation from CLIP~\cite{radford2021learning} to the discriminator, ultimately improving the generation performance. Note we choose Alexnet
model with linear configuration for LPIPS evaluation.

\textbf{Generalization of KD-DLGAN:} 
We study the generalization of our KD-DLGAN by performing experiments with different GAN architectures, generation tasks and the number of training samples. Specifically, as shown in Table~\ref{Low Shot Generation}-\ref{tab:ImageNet-biggan}, we perform extensive evaluations over BigGAN and StyleGAN-v2. Meanwhile, we benckmark KD-DLGAN on object generation tasks with CIFAR and ImageNet and face generation tasks with 100-shot and AFHQ. Besides, we perform extensive evaluations on 100-shot and AFHQ with few hundred samples, 
CIFAR with 100\%, 20\% and 10\% data, ImageNet with 10\%, 5\% and 2.5\% data.

\textbf{Comparison with state-of-the-art knowledge distillation methods:}
KD-DLGAN is the first to explore the idea of knowledge distillation in data-limited image generation.
To validate the superiority of our designs, we compare KD-DLGAN with state-of-the-art knowledge distillation methods designed for other tasks in Table~\ref{tab:other knowledge distill}. It shows that our KD-DLGAN outperforms the state-of-the-art knowledge distillation approaches consistently by large margins. The superior generation performance demonstrates the effectiveness of our designed generative knowledge distillation techniques for data-limited image generation. 

\textbf{Comparison with other Vision-Language teacher models:} KD-DLGAN adopted CLIP~\cite{radford2021learning} as the teacher model for knowledge distillation. We perform experiments to study how different vision-language models affect the generation performance. As shown in Table~\ref{tab:other vision language}, different vision-language models produce quite similar FIDs. We conjecture that these  pertrained models provide sufficient vision-language information for distillation and the performance gain mainly comes from our designed generative knowledge distillation techniques instead of the selected teacher model.

\begin{table}[t]
\centering
\resizebox{0.85\columnwidth}{!}{
\begin{tabular}{l|cc}
\toprule
\multirow{2}*{Method} & \multicolumn{1}{c}{CIFAR-10} &  \multicolumn{1}{c}{CIFAR-100} \\
& \multicolumn{1}{c}{10\% data} & \multicolumn{1}{c}{10\% data}  \\
\midrule
DA~\cite{zhao2020differentiable} (Baseline) &0.202  & 0.236 \\ 
KD-DLGAN w/o CKGD & 0.204 & 0.237 \\
KD-DLGAN & \textbf{0.221} & \textbf{0.264} \\
\bottomrule
\end{tabular}}
\caption{
KD-DLGAN improves the generation diversity clearly. And the improvement is largely attributed to CGKD, which distills diverse image-text correlations from CLIP~\cite{radford2021learning} to the discriminator. We report LPIP ($\uparrow$) averaged over three runs. 
}
\label{tab:LPIP}
\end{table}

\begin{table}[t]
\centering
\resizebox{0.85\columnwidth}{!}{
\begin{tabular}{l|cc}
\toprule
\multirow{2}*{Method} & \multicolumn{1}{c}{CIFAR-10} &  \multicolumn{1}{c}{CIFAR-100}  \\
& \multicolumn{1}{c}{10\% data} & \multicolumn{1}{c}{10\% data}  \\
 \midrule
DA~\cite{zhao2020differentiable} (Baseline) &  23.34 \textpm \ 0.09 & 35.39 \textpm \ 0.08\\ 
Fitnets~\cite{romero2014fitnets}&  22.03 \textpm \ 0.07 & 33.93 \textpm \ 0.09\\
Label Distillation\cite{hinton2015distilling} & 20.46 \textpm \ 0.10 & 34.14 \textpm \ 0.11 \\
PKD~\cite{passalis2018learning} & 21.34 \textpm \ 0.08 & 32.15 \textpm \ 0.13\\
SPKD~\cite{tung2019similarity} &  19.11 \textpm \ 0.07 & 31.97 \textpm \ 0.10 \\
KD-DLGAN (Ours) &\textbf{14.20} \textpm \ 0.06 & \textbf{18.03} \textpm \ 0.11\\
\bottomrule
\end{tabular}
}
\caption{
KD-DLGAN  outperforms the state-of-the-art knowledge distillation methods by large margins, demonstrating the effectiveness of the two generative knowledge distillation techniques designed specifically for data-limited image generation. We report FID($\downarrow$) averaged over three runs.
}
\label{tab:other knowledge distill}
\end{table}

\begin{table}[t]
\centering
\resizebox{0.8\columnwidth}{!}{
\begin{tabular}{l|cc}
\toprule
\multirow{2}*{Method} & \multicolumn{1}{c}{CIFAR-10} & \multicolumn{1}{c}{CIFAR-100} \\
 & \multicolumn{1}{c}{10\% data} & \multicolumn{1}{c}{10\% data}  \\
 \midrule
 DA~\cite{zhao2020differentiable} (Baseline) &   23.34 \textpm \ 0.09 & 35.39 \textpm \ 0.08\\ 
+ TCL~\cite{yang2022vision} & 14.98 \textpm \ 0.09 & 18.43 \textpm \ 0.12 \\
+ BLIP~\cite{li2022blip} & 15.74 \textpm \ 0.10 & 18.88 \textpm \ 0.11\\
+ CLIP~\cite{radford2021learning}  (Ours) &  \textbf{14.20} \textpm \ 0.06 & \textbf{18.03} \textpm \ 0.11\\
 \bottomrule
\end{tabular}
}
\caption{Employing 
different pretrained vision-language models as teacher models, the results are similar. We report FID($\downarrow$) averaged over three runs.
}
\label{tab:other vision language}
\end{table}

\begin{table}[t]
\centering
\resizebox{0.85\columnwidth}{!}{
\begin{tabular}{l|cc}
\toprule
\multirow{2}*{Method} & \multicolumn{1}{c}{CIFAR-10} &  \multicolumn{1}{c}{CIFAR-100} \\
 & \multicolumn{1}{c}{10\% data} & \multicolumn{1}{c}{10\% data}  \\
 \midrule
 DA~\cite{zhao2020differentiable} (Baseline) & 23.34 \textpm \ 0.09 & 35.39 \textpm \ 0.08\\ 
Vision-aided GAN~\cite{Kumari_2022_CVPR} &  16.24  \textpm \ 0.08 & 19.11 \textpm \ 0.10\\
 KD-DLGAN (Ours)  & \textbf{14.20} \textpm \ 0.06 & \textbf{18.03} \textpm \ 0.11\\
 \bottomrule
\end{tabular}
}
\caption{
KD-DLGAN outperforms CLIP-based Vision-add GAN~\cite{Kumari_2022_CVPR}, demonstrating the effectiveness of our KD-DLGAN in mitigating discriminator overfitting and improving the generation performance. We report FID($\downarrow$) averaged over three runs.
}
\label{tab:otherclip}
\end{table}

\textbf{Comparison with other CLIP-based methods:} KD-DLGAN distills knowledge from CLIP~\cite{radford2021learning} to the discriminator with two novelly designed generative knowledge distillation techniques while Vision-aided GAN~\cite{Kumari_2022_CVPR} employs off-the-shelf models as additional discriminators for data-limited generation. Table~\ref{tab:otherclip} compares KD-DLGAN with Vision-aided GAN. We can observe that KD-DLGAN outperforms CLIP-based Vision-aided GAN consistently, demonstrating its effectiveness in mitigating discriminator overfitting and improving generation performance.

\section{Conclusion}

In this paper, we present KD-DLGAN, a novel data-limited image generation framework that introduces knowledge distillation for effective GAN training with limited data. We design two novel generative knowledge distillation techniques, including aggregated generative knowledge distillation (AGKD) and correlated generative knowledge distillation (CGKD).  AGKD mitigates the discriminator overfitting by forcing harder learning tasks and distilling more general knowledge from CLIP.  CGKD improves the generation diversity by distilling and preserving the diverse image-text correlation within CLIP. Extensive experiments show that both AGKD and CGKD can improve the generation performance and combining them leads to the best performance. We also show that KD-DLGAN complements the state-of-the-art data-limited generation methods consistently. Moving forward, we will explore KD-DLGAN in more tasks such as image translation and editing. 

\section*{Acknowledgement}
This study is funded by the Ministry of Education Singapore, under the Tier-1 scheme with a project number RG94/20, as well as cash and in-kind contribution from Singapore Telecommunications Limited (Singtel), through Singtel Cognitive and Artificial Intelligence Lab for Enterprises (SCALE@NTU).

{\small
\bibliographystyle{ieee_fullname}
\bibliography{egbib}
}

\end{document}